\documentclass[conference, 10pt, letter]{IEEEtran}
\IEEEoverridecommandlockouts
\usepackage[noend]{algpseudocode}

\usepackage{algorithm}
\usepackage{graphicx}                                      
\usepackage{amsmath}
\usepackage{amssymb}
\usepackage{amsfonts}
\usepackage{amsthm}
\usepackage[mathcal]{eucal}
\usepackage{booktabs}
\usepackage{enumerate}
\usepackage{multirow}
\usepackage[subrefformat=parens,farskip=0pt,justification=centering]{subfig}
\usepackage{color}
\usepackage{cite}                                          
\usepackage{comment}                                       
\usepackage{soul}                                          
\soulregister\cite7
\soulregister\ref7
\soulregister\pageref7
\usepackage{etoolbox}                                      
\usepackage{url}
\usepackage{nth}                                           
\usepackage{bm}                                            
\usepackage{courier}
\usepackage{balance}
\usepackage{threeparttable}
\usepackage[bookmarks=false]{hyperref}
\hypersetup{
    colorlinks = true,
    citecolor  = blue,
    linkcolor  = blue,
    urlcolor   = blue,
}
\usepackage{tikz}
\usetikzlibrary{positioning,calc,fit,decorations.pathmorphing}
\usepackage{filecontents}                                  
\usepackage{pgfplots}
\usepackage{pgfplotstable}
\pgfplotsset{compat=newest}
\usepackage{caption}
\usepackage{cleveref}
\Crefformat{figure}{Fig.~#2#1#3}                           
\Crefname{subfigure}{Fig.}{Figs.}
\Crefname{figure}{Fig.}{Figs.}
\Crefformat{table}{TABLE~#2#1#3}                           
\definecolor{CUHKorange}{RGB}{244,106,18} 
\definecolor{CUHKblue}{RGB}{0,111,190}    
\definecolor{CUHKgreen}{RGB}{0,127,128}   
\definecolor{CUHKred}{RGB}{228,46,36}     
\definecolor{CUHKyellow}{RGB}{198,148,34} 
\definecolor{CUHKdark}{RGB}{114,44,114}   
\definecolor{CUHKmiddle}{RGB}{144,44,144} 
\definecolor{CUHKlight}{RGB}{167,44,167} 

\algrenewcommand\textproc{\texttt}

\makeatletter
\let\OldStatex\Statex
\renewcommand{\Statex}[1][3]{%
  \setlength\@tempdima{\algorithmicindent}%
  \OldStatex\hskip\dimexpr#1\@tempdima\relax
}
\makeatother

\RequirePackage[normalem]{ulem} 
\RequirePackage{color}\definecolor{RED}{rgb}{1,0,0}\definecolor{BLUE}{rgb}{0,0,1} 

\usepackage[frozencache,cachedir=.]{minted}
\usepackage{adjustbox}
\usepackage{cite}
\usepackage{tcolorbox}
\usepackage{amsmath,amssymb,amsfonts}
\usepackage{algorithm}
\usepackage{algpseudocode}
\usepackage{graphicx}
\usepackage{textcomp}
\usepackage{xcolor}
\usepackage{cleveref}
\usepackage{comment}
\usepackage{bbm}
\usepackage{booktabs}
\usepackage{multirow}
\usepackage{array}
\usepackage{caption}
\usepackage{subfig}
\definecolor{NVgreen}{RGB}{118,185,0}
\definecolor{NVblack}{RGB}{0,0,0}
\definecolor{NVlgrey}{RGB}{205,205,205}
\definecolor{NVmgrey}{RGB}{140,140,140}
\definecolor{NVdgrey}{RGB}{94,94,94}

\definecolor{NVemerald}{RGB}{0,133,100}
\definecolor{NVamethyst}{RGB}{93,22,130}
\definecolor{NVintel}{RGB}{0,113,197}
\definecolor{NVgarnet}{RGB}{137,12,88}
\definecolor{NVfluorite}{RGB}{250,194,0}

\begin{document}
\bstctlcite{IEEEexample:BSTcontrol}
\graphicspath{{figs/}}

\title{VerilogEval: Evaluating Large Language Models for Verilog Code Generation}
\author{\IEEEauthorblockN{Mingjie Liu, Nathaniel Pinckney, Brucek Khailany, and Haoxing Ren}
\IEEEauthorblockA{{NVIDIA Corporation} \\
\{mingjiel, npinckney, bkhailany, haoxingr\}@nvidia.com 
}
}
\maketitle
\begin{abstract}
The increasing popularity of large language models (LLMs) has paved the way for their application in diverse domains.
This paper proposes a benchmarking framework tailored specifically for evaluating LLM performance in the context of Verilog code generation for hardware design and verification. 
We present a comprehensive evaluation dataset consisting of 156 problems from the Verilog instructional website HDLBits. 
The evaluation set consists of a diverse set of Verilog code generation tasks, ranging from simple combinational circuits to complex finite state machines.
The Verilog code completions can be automatically tested for functional correctness by comparing the transient simulation outputs of the generated design with a golden solution. 
We also demonstrate that the Verilog code generation capability of pretrained language models could be improved with supervised fine-tuning by bootstrapping with LLM generated synthetic problem-code pairs. 
\end{abstract}

\section{Introduction}
The escalating popularity of Large Language Models (LLMs), characterized by their remarkable capacity to comprehend and generate human-like text, has opened up a realm of possibilities across diverse domains~\cite{LLM_fewshot_learners, gpt4, general_AI}. 
LLMs tailored for specific domains have garnered significant attention owing to their impressive performance on both general-purpose benchmarks and specialized tasks within domains like financial engineering~\cite{bloomberggpt}, biomedical studies~\cite{biomegatron, BioMedLM}, and general scientific research~\cite{galactica}.
When it comes to coding, LLMs can assist developers by suggesting code snippets, offering solutions to common programming challenges, and even explaining complex concepts in a more accessible manner~\cite{copilot,codegen}.

In the realm of Electronic Design Automation, LLMs provide the potential to aid engineers in designing and verifying digital systems, providing insights into Verilog coding, optimizing circuits, and automating time-consuming tasks~\cite{chipchat,chipgpt}. 
A number of studies have initiated the evaluation of LLMs' potential in generating Verilog code.
Thakur et al.~\cite{thakur2023benchmarking} fine-tuned CodeGen~\cite{codegen} models which was evaluated on 17 designs.
A later follow up work~\cite{chipchat} further demonstrate the ability to design chip-level circuits with ChatGPT.
RTLLM~\cite{RTLLM} propose a benchmark framework with 30 designs, which focus on increasing the benchmark design scalability.
The authors further improved the solution quality with simple and effective prompt engineering techniques.

\begin{figure}
    \centering
    \includegraphics[width=0.98\linewidth]{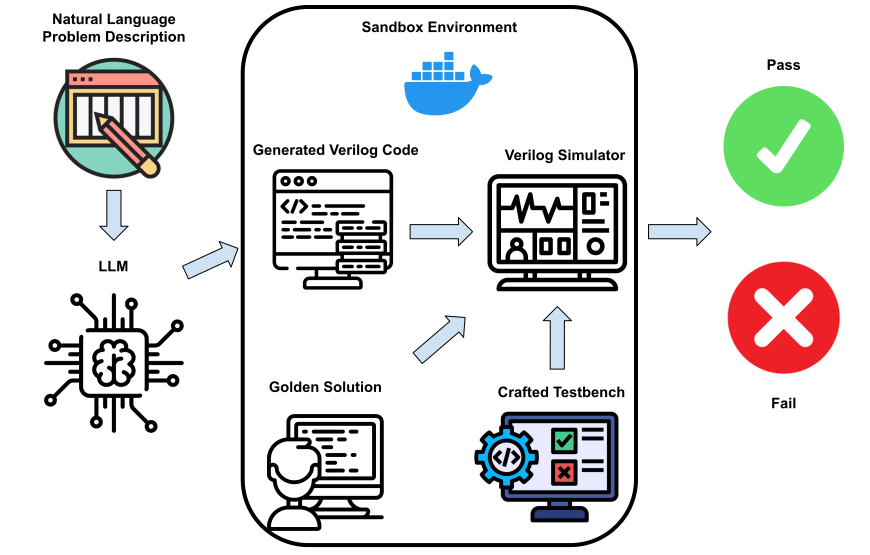}
    \caption{\textbf{VerilogEval} uses a sandbox environment for simple and reproducible evaluation of LLM Verilog code generation}
    \label{fig:verilog_eval}
\end{figure}

While LLMs have proven to be powerful tools, their pretraining phase, characterized by unsupervised training loss, often lacks alignment with specific tasks.
To enhance performance, supervised fine-tuning (SFT) is used~\cite{flan}, involving task-specific data to adapt to requirements. 
Ensuring model alignment is imperative for achieving improved performance, driving the investigation of increasingly computationally demanding techniques, such as reinforcement learning with human feedback~\cite{rlhf, rlhf_anthropic}.
The cost associated with acquiring labeled data also remains a barrier, prompting a growing interest in alternative annotation-free alignment techniques. 
Self-Instruct~\cite{self_instruct} starts with a seed task, using LLMs to create more instructions and instances. 
WizardLM's Evol-Instruct~\cite{wizardlm} evolves instructions for a diverse dataset, which is further applied to code generation~\cite{wizardcoder}. 
Additionally, a recent study utilized GPT-4 to generate a high-quality synthetic textbook dataset, achieving superior coding capabilities at 1/100th of the cost of other models~\cite{textbook_all_you_need}.
Within the realm of Verilog coding, there remains a notable gap in the exploration of supervised fine-tuning for model enhancement.

Moreover, notwithstanding commendable endeavors, recent research in Verilog code benchmarking has revealed limitations concerning its comprehensiveness, quantity, and the diversity of problems studied.
Effective benchmarks should exhibit diversity, encompassing a wide array of topics, to mitigate testing variance. 
Furthermore, they should offer unambiguous problem descriptions, ensuring that solutions can be assessed with clear distinctions regarding correctness. 
In addition, reliability and automation are key factors, enabling the straightforward evaluation of generated code through robust testing procedures. 

Our research addresses these gaps through the introduction of \textbf{VerilogEval}\footnote{\url{https://github.com/NVlabs/verilog-eval}}, a open-source benchmark that encompasses a diverse array of questions, offers clear and unambiguous problem descriptions, and incorporates automated, easily reproducible testing procedures. 
This contribution significantly enhances the robustness and effectiveness of the evaluation framework for Verilog code generation and assessment.
Our specific contributions are as follows:
\begin{itemize}
    \item We present a comprehensive evaluation dataset comprising 156 problems sourced from the HDLBits. These problems have undergone meticulous curation, ensuring both clarity and diversity. 
    \item We developed a benchmarking framework wherein Verilog code completions are subjected to automatic functional correctness testing. 
    \item We constructed a synthetic supervised fine-tuning dataset by leveraging LLMs to generate problem descriptions paired with Verilog code. This dataset is employed in extensive experiments on SFT, further enhancing the model's proficiency in Verilog coding tasks.
\end{itemize}


\section{Evaluation Framework}

In this section we discuss the details of our evaluation framework and evaluation dataset collection.
Our work closely follows the widely adopted Python coding benchmark HumanEval~\cite{codex} for best practices.
\textbf{VerilogEval} is presented in~\Cref{fig:verilog_eval} where we develop a sandbox environment for simple and reproducible evaluation of LLM Verilog code generation.

\begin{figure}[h]
\scriptsize	
\begin{center}
\begin{tcolorbox}[width=0.9\linewidth, halign=left, colframe=red, colback=blue!30, boxsep=0.01mm, arc=1.5mm, left=2mm, right=2mm, boxrule=0.5pt]
\textbf{System Prompt (Optional):}\\
You only complete chats with syntax correct Verilog code. End the Verilog module code completion with 'endmodule'. Do not include module, input and output definitions.
\end{tcolorbox}
\vspace{-0.116in}
\begin{tcolorbox}[width=0.9\linewidth, halign=left, colframe=red, colback=green!30, boxsep=0.01mm, arc=1.5mm, left=2mm, right=2mm, boxrule=0.5pt]
\textbf{Question Prompt:}\\ 
Implement the Verilog module based on the following description. Assume that signals are positive clock/clk edge triggered unless otherwise stated.
\end{tcolorbox}
\vspace{-0.116in}
\begin{tcolorbox}[width=0.9\linewidth, halign=left, colframe=red, colback=yellow!30, boxsep=0.01mm, arc=1.5mm, left=2mm, right=2mm, boxrule=0.5pt]
\textbf{Problem Description:}\\
Given an 8-bit input vector [7:0], reverse its bit ordering.
\begin{minted}[breaklines]{verilog}
module top_module (
	input [7:0] in,
	output [7:0] out
);
\end{minted}
\end{tcolorbox}
\begin{tcolorbox}[width=0.9\linewidth, halign=left, colframe=red, colback=red!30, boxsep=0.01mm, arc=1.5mm, left=2mm, right=2mm, boxrule=0.5pt]
\textbf{Canonical Solution:}\\
\begin{minted}[breaklines]{verilog}
    assign {out[0],out[1],out[2],out[3],out[4], out[5],out[6],out[7]} = in;	
endmodule
\end{minted}
\end{tcolorbox}
\end{center}
\caption{Example of \texttt{vectorr} in \textbf{VerilogEval-human}. The \textbf{Problem Description} includes both natural language description and module header, input, and output definition.}
\label{fig:question_example}
\end{figure}

\subsection{\textbf{VerilogEval} Evaluation Set}
We evaluate functional correctness on a selected set of problems from the Verilog instructional website HDLBits\footnote{\url{https://hdlbits.01xz.net/wiki/Problem_sets}}.
HDLBits is a collection of digital circuit design exercises and an online judge for learning digital logic using the Verilog hardware description language.
The evalutation set consists of diverse Verilog coding tasks, ranging from module implementation of simple combinational circuits to complex finite state machines, code debugging, and testbench construction.

We focus on generating \textit{self-contained}\footnote{Example of a removed question that is not \textit{self-contained}: \url{https://hdlbits.01xz.net/wiki/Module_cseladd}.} Verilog modules from natural language text descriptions.
We define a Verilog module as \textit{self-contained} if the module implementation does not require instantiation of any other modules.
We emphasize the significance of module instantiation as a crucial capability in Verilog, playing an essential role in constructing extensive system-level designs. 
It's important to note that our evaluation does not delve into this topic. 
However, while most problems in \textbf{VerilogEval} are intentionally concise, they demand the LLM to possess a comprehension of hardware design along with adept problem-solving skills in areas encompassing circuits, Boolean logic, state transitions, and more.

\Cref{fig:question_example} shows an example of the problem \texttt{vectorr}. 
\textbf{Problem Description} includes both natural language description and module header and IO definition.
Including the module header removes ambiguity such as the bit width of signals.
The \textbf{Question Prompt} is concatenated with \textbf{Problem Description} and sent to the LLM for inference.
\textbf{Canonical Solution} is provided as the golden solution for testing.

\subsection{Problem Descriptions}
Although HDLBits serves as a valuable resource for Verilog coding challenges, a significant portion of the website's problem descriptions are not readily compatible with text-only language models. 
These problem descriptions rely on various modalities, frequently incorporating circuit schematic images, state transition diagram graphs, Boolean logic tables, and Karnaugh maps.
We explore with two methods for generating text-only problem descriptions for these problem sets.

\begin{figure}[ht]
\scriptsize	
\begin{center}
\begin{tcolorbox}[width=0.9\linewidth, halign=left, colframe=red, colback=blue!30, boxsep=0.01mm, arc=1.5mm, left=2mm, right=2mm, boxrule=0.5pt]
\textbf{System Prompt:}\\
Explain the high-level functionality of the Verilog module. Use as many high-level concepts that are directly applicable to describe the code, say at the level of an undergraduate EECS major, but do not include extraneous details that aren’t immediately applicable. Use text-based truth tables and state transition graphs when necessary. Speak concisely as if this was a specification for a circuit designer to implement. You should only reply with descriptive natural language and not use any code.
\end{tcolorbox}
\vspace{-0.116in}
\begin{tcolorbox}[width=0.9\linewidth, halign=left, colframe=red, colback=green!30, boxsep=0.01mm, arc=1.5mm, left=2mm, right=2mm, boxrule=0.5pt]
\textbf{Few-shot Examples (Optional):}\\ 
Question: Explain the high-level functionality of the Verilog module. 
\begin{minted}[breaklines]{verilog}
module top_module (
	input [7:0] in,
	output [7:0] out
);
    assign {out[0],out[1],out[2],out[3],out[4], out[5],out[6],out[7]} = in;	
endmodule
\end{minted}
Answer: This top Verilog module is a 8-bit unidirectional data port. It takes an 8-bit input in, and outputs an 8-bit signal out. The output of the module is assigned to out based on the value of in. If in = {a,b,c,d,e,f,g,h}, then out = {h,g,f,e,d,c,b,a}.
\end{tcolorbox}
\vspace{-0.116in}
\begin{tcolorbox}[width=0.9\linewidth, halign=left, colframe=red, colback=yellow!30, boxsep=0.01mm, arc=1.5mm, left=2mm, right=2mm, boxrule=0.5pt]
\textbf{User Prompt:}\\
Question: Explain the high-level functionality of the Verilog module. 
\begin{minted}[breaklines]{verilog}
module top_module (
	input [5:0] y,
	input w,
	output Y1,
	output Y3
);
    assign Y1 = y[0]&w;
    assign Y3 = (y[1]|y[2]|y[4]|y[5]) & ~w;
endmodule
\end{minted}
\end{tcolorbox}
\begin{tcolorbox}[width=0.9\linewidth, halign=left, colframe=red, colback=red!30, boxsep=0.01mm, arc=1.5mm, left=2mm, right=2mm, boxrule=0.5pt]
\textbf{GPT-3.5 Response:}\\
Answer: This Verilog module is a combinational logic circuit that takes in 6-bit input vector y and a single bit input w. The module then outputs two signals Y1 and Y3. Y1 is the logical AND of the first bit of the input vector y and the single bit input w. Y3 is the logical AND of the logical OR of the second, third, fifth and sixth bits of the input vector y and the logical NOT of the single bit input w. In other words, Y1 is true if both the first bit of the input vector y and the single bit input w are true, and Y3 is true if at least one of the second, third, fifth and sixth bits of the input vector y is true and the single bit input w is false.
\end{tcolorbox}
\end{center}
\caption{\textbf{VerilogEval-machine} use \texttt{gpt-3.5-turbo} to generate problem descriptions for \texttt{2012\_q2b}.}\vspace{-0.1in}
\label{fig:machine_description}
\end{figure}

\subsubsection{\textbf{VerilogEval-machine}}
We completely disregard the descriptions on the website and opt to utilize LLMs for the automated creation of problem descriptions. 
We employ the prompt template depicted in \Cref{fig:machine_description}, employing \texttt{gpt-3.5-turbo}.
Initially, we create all problem descriptions using zero-shot methods. 
We validate these descriptions by using the LLM to produce code solutions. 
Problem descriptions are considered invalid if none of the generated completions succeed across 100 samples, and such descriptions are then discarded.
Surprisingly, among the pool of 156 candidate problems, 108 of them yield successful solutions upon initial sampling.
Subsequent to this, we consider the valid generated descriptions as few-shot examples (4-shot) and proceed to further sample unsolved problems. 
In this phase, we iteratively sample descriptions along with their corresponding code completions (8 completions per description). 
Descriptions are labeled as valid as soon as any of the corresponding code completions pass testing.
Sampling for each problem is halted upon reaching a allocated sampling budget, resulting in an increase of 35 additional solutions.
In total we generated 143 valid problem descriptions.

\begin{figure}[h]
\scriptsize	
\begin{center}
\begin{tcolorbox}[width=0.9\linewidth, halign=left, colframe=red, colback=yellow!30, boxsep=0.01mm, arc=1.5mm, left=2mm, right=2mm, boxrule=0.5pt]
\textbf{User Prompt:}\\
I want you to write Verilog from a state transition diagram. How can I give you this diagram in a text format?
\end{tcolorbox}
\begin{tcolorbox}[width=0.9\linewidth, halign=left, colframe=red, colback=red!30, boxsep=0.01mm, arc=1.5mm, left=2mm, right=2mm, boxrule=0.5pt]
\textbf{GPT-3.5 Response:}\\
For a more graphical representation using text, you can specify the states with their output values and the transitions like this:
\begin{minted}[breaklines]{python}
StateA (0) --0--> StateA
StateA (0) --1--> StateB
StateB (1) --0--> StateC
StateB (1) --1--> StateB
StateC (0) --0--> StateA
StateC (1) --1--> StateB
\end{minted}
\end{tcolorbox}
\end{center}
\caption{ChatGPT guidance on state transition diagrams.}
\label{fig:state_graphs}
\end{figure}

\subsubsection{\textbf{VerilogEval-human}}
\begin{figure}[ht]
  \scriptsize
  \centering
    \begin{tcolorbox}[width=0.9\linewidth, halign=left, colframe=red, colback=white, boxsep=0.01mm, arc=1.5mm, left=2mm, right=2mm, boxrule=0.5pt]
    Consider the state machine shown below:
    \begin{minted}[breaklines]{python}
     A (0) --1--> B
     A (0) --0--> A
     B (0) --1--> C
     B (0) --0--> D
     C (0) --1--> E
     C (0) --0--> D
     D (0) --1--> F
     D (0) --0--> A
     E (1) --1--> E
     E (1) --0--> D
     F (1) --1--> C
     F (1) --0--> D
    \end{minted}
    Assume that a one-hot code is used with the state assignment y[5:0] = 000001(A), 000010(B), 000100(C), 001000(D), 010000(E), 100000(F). \\
    Write a Verilog for the signal Y1, which is the input of state flip-flop y[1], for the signal Y3, which is the input of state flip-flop y[3]. Derive the Verilog by inspection assuming a one-hot encoding.
    \end{tcolorbox}
    \vspace{-1.95in}
    \hspace*{\fill}\includegraphics[width=0.16\textwidth]{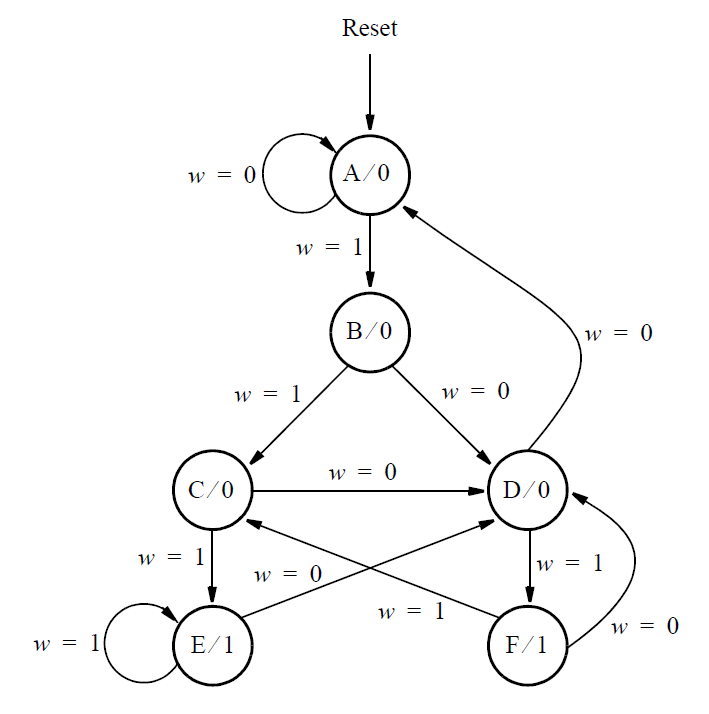}\hspace*{0.8in}
    
    \vspace{0.65in}
    \vspace{0.4cm}
    {\small (a) \texttt{2012\_q2b}}
    \vspace{0.2cm}

    \begin{tcolorbox}[width=0.9\linewidth, halign=left, colframe=red, colback=white, boxsep=0.01mm, arc=1.5mm, left=2mm, right=2mm, boxrule=0.5pt]
    Implement the circuit described by the Karnaugh map below.
    \begin{minted}[breaklines]{python}
    
          ab
   cd   00 01 11 10
    00 | 0 | 1 | 0 | 1 |
    01 | 1 | 0 | 1 | 0 |
    11 | 0 | 1 | 0 | 1 |
    10 | 1 | 0 | 1 | 0 |
    \end{minted}
    \end{tcolorbox}
    \vspace{-0.845in}
    \hspace*{\fill}\includegraphics[width=0.1\textwidth]{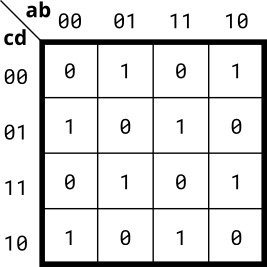}\hspace*{1.0in}

    \vspace{0.4cm}
    {\small (b) \texttt{kmap4}}
    \vspace{0.2cm}

    \begin{tcolorbox}[width=0.9\linewidth, halign=left, colframe=red, colback=white, boxsep=0.01mm, arc=1.5mm, left=2mm, right=2mm, boxrule=0.5pt]
    Given the finite state machine circuit described below, assume that the D flip-flops are initially reset to zero before the machine begins. \\
    Build this circuit in Verilog. \\
    \texttt{Input x goes to three different two-input gates: a XOR, an AND, and a OR gate. Each of the three gates is connected to the input of a D flip-flop and then the flip-flop outputs all go to a three-input XNOR, whose output is Z. The second input of the XOR is its corresponding flip-flop's output, the second input of the AND is its corresponding flip-flop's complemented output, and finally the second input of the OR is its corresponding flip-flop's complementary output.}
    \begin{center}
        \includegraphics[width=0.6\textwidth]{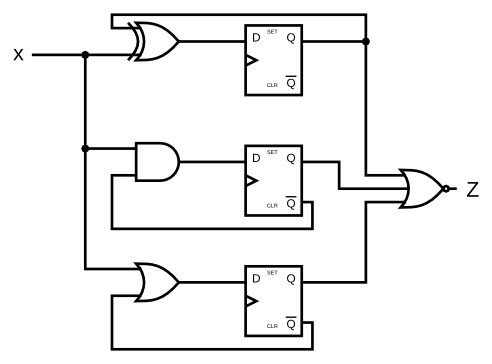}
    \end{center}
    \end{tcolorbox}
    
    \vspace{0.2cm}
    {\small (c) \texttt{ece241\_2014\_q4}}
    
    \caption{Examples of \textbf{VerilogEval-human} descriptions. We show original website descriptions alongside manually converted text format.}
    \label{fig:conversion_example}
\end{figure}

We engaged in manual review and conversion of problem descriptions from the website into a text-only structure. 
We dedicated particular attention to addressing ambiguity within the problem descriptions, particularly when precisely determining attributes such as the clock's posedge or negedge triggering, whether reset and enable signals are active high or active low, and whether they operate synchronously or asynchronously.
Boolean logic tables and Karnaugh maps were transformed into textual tabular formats. Circuit schematic diagrams were translated into natural language explanations of the connections between logical gates.
For sequential waveforms, we meticulously detailed all signal values at each transition edge of the clock, presented in a tabular layout with an added column for time steps.
One particular challenge we confronted revolved around the task of converting state transition graphs into a text-based representation. 
To tackle this, we turned to ChatGPT for guidance, as depicted in~\Cref{fig:state_graphs}. 
We ultimately adopted the edge list-based format to depict these state transition graphs.
Examples of manually converted descriptions are shown in~\Cref{fig:conversion_example}.
Initial explorations were conducted regarding Verilog code completion by employing the converted formats. 
Notably, ChatGPT exhibited the capability to generate meaningful code using these formats for simple problems.
We manually converted 156 problem descriptions in total.

Comparing the descriptions between \textbf{machine} and \textbf{human}, we find that \textbf{machine} descriptions are often more verbose (\texttt{vectorr} in~\Cref{fig:question_example,fig:machine_description}).
Although the model is directed to generate high-level explanations, produced \textbf{machine} descriptions frequently delve into low-level details. 
These descriptions tend to mirror the code's implementation line by line, rather than focusing on the overarching functionality of the circuit (\texttt{2012\_q2b} in~\Cref{fig:machine_description,fig:conversion_example}).
Furthermore, despite that we have taken steps to ensure that all \textbf{machine} descriptions are capable of producing passing solutions through LLMs, we cannot guarantee the absence of ambiguity and errors.
Nevertheless, \textbf{VerilogEval-machine} remains a valuable benchmark, particularly for assessing LLM's competence in comprehending low-level instructions and generating syntactically and functionally accurate Verilog code.

\begin{table}[ht]
\centering
\begin{tabular}{@{}cccc@{}}
\toprule[1pt]\midrule[0.3pt]
Benchmark           & Language & Train & Test  \\ \midrule
HumanEval~\cite{codex}           & Python   & -     & 164   \\
MBPP~\cite{mbpp}                & Python   & 374   & 500   \\
APPS~\cite{apps}                & Python   & 5,000 & 5,000 \\ \midrule
\textbf{VerilogEval-machine} & Verilog  & 8,502 & 143   \\
\textbf{VerilogEval-human}   & Verilog  & -     & 156   \\ \midrule[0.3pt] \bottomrule[1pt]
\end{tabular}
\caption{\textbf{VerilogEval} in comparison with popular (python) coding benchmarks.}
\label{tab:comparison}
\end{table}

~\Cref{tab:comparison} summarizes our \textbf{VerilogEval} benchmark and compares with existing popular coding benchmarks in python.
We further detail our method of generating synthetic supervised fine-tuning dataset in~\Cref{sec:synthetic_data}.

\subsection{Automated Testing Environment}
Unlike software programming languages like Python, the behavior of Verilog modules are typically evaluated through simulations.
To enable automated testing, we compare simulation results between generated code completions with golden reference solutions.
We assert for output signal correctness at clock (posedge and/or negedge) transition edges for sequential circuits, while for combinational circuits, we validate them when any input signals changes.
Our testbench incorporates two categories of input signals for each problem: manually crafted test patterns of significance, and randomly generated test patterns.
Randomly generated test patterns may span from a few hundred clock cycles for simple problems to several thousand cycles for more complex ones.

We adapted the sandbox environment to safetly run untrusted programs from HumanEval~\cite{codex}.
We built and installed the open-source ICARUS Verilog~\cite{iverilog} simulator in a docker container.
We note that our evaluation of Verilog syntax is limited by the simulator, which might not include all features of Verilog HDL IEEE-1364 standard.
Simulation and testing are handled under the hood and results can be produced using just a single line of command.

\subsection{Evaluation Metric}
\begin{figure}[h]
  \centering
    \subfloat{\includegraphics[width=0.24\textwidth]{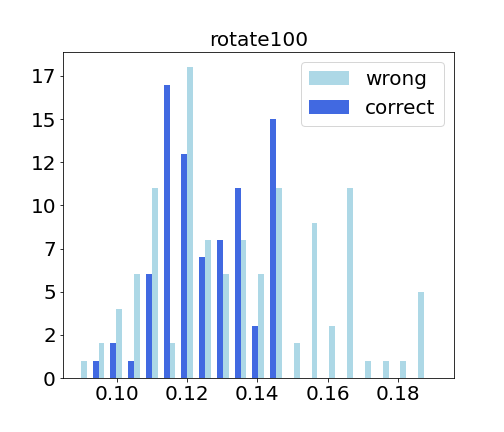}}
    \subfloat{\includegraphics[width=0.24\textwidth]{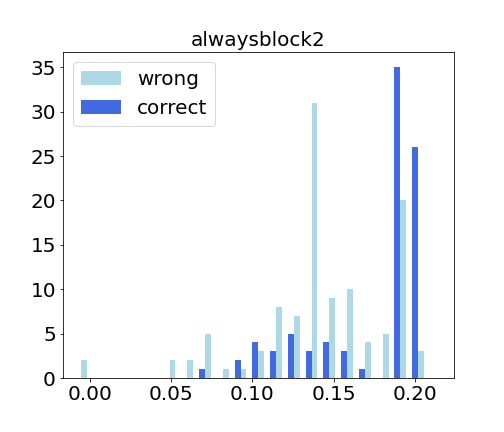}}
    \caption{BLEU score probability densities for correct and wrong solutions from \texttt{codegen-16B-verilog}~\cite{thakur2023benchmarking} for 2 tasks from \textbf{VerilogEval-human}.}
    \label{fig:bleu_compare}
\end{figure}

Early works on code evaluations report on match-based metrics such as BLEU score~\cite{bleu}. 
However, recent works~\cite{codexglue, codex} have argued that such metrics does not correlate well with functional correctness.
In~\Cref{fig:bleu_compare} we show that Verilog coding exhibit similar issues, where the distributions of correct versus wrong solutions are not clearly seperable based on BLEU score probability densities.

\begin{figure}[h]
    \centering
    \includegraphics[width=0.9\linewidth]{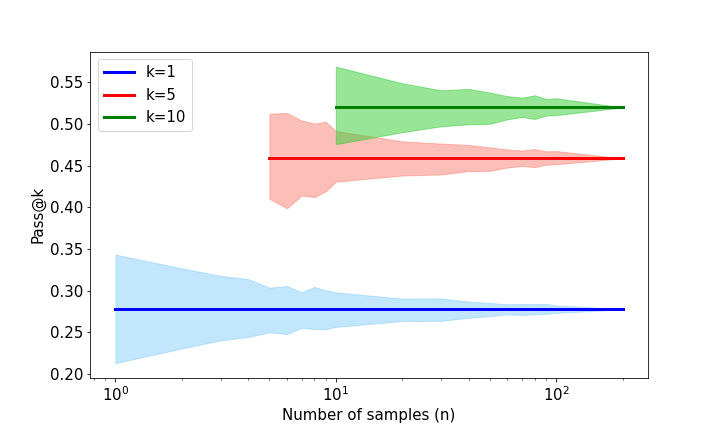}
    \caption{Variance in estimating pass@\textit{k} with $n$. Samples from \texttt{codegen-16B-verilog}~\cite{thakur2023benchmarking} for \textbf{VerilogEval-human}.}
    \label{fig:variance}
\end{figure}

We follow recent work in directly measuring code functional correctness through pass@\textit{k} metric~\cite{codex, kulal2019spoc, mbpp}, where a problem is considered solved if any of the \textit{k} samples passes the unit tests.
We also suggest using the unbiased estimator from~\cite{codex}:
\begin{equation}
    pass@\textit{k}:=\displaystyle \mathop{\mathbb{E}}_{Problems}\left[1-  \frac{\binom{n-c}{k}} { \binom{n}{k}}  \right],
\label{eq:pass_k}
\end{equation}
where we generate $n\geq k$ samples per task in which $c\leq n$ samples pass testing.
In~\Cref{fig:variance} we show that the number of samples $n$ need to be sufficiently large to produce low variance estimates for pass@\textit{k}.

\section{Supervised Fine-Tuning}
This section provides our findings concerning the supervised fine-tuning (SFT) of Large Language Models (LLM) for Verilog coding. 
We elucidate our approach to the generation of synthetic SFT data, achieved by utilizing LLMs to create problem descriptions, detailed in~\Cref{sec:synthetic_data}. 
Subsequently, \Cref{sec:sft_results} comprises a comprehensive suite of supervised fine-tuning (SFT) experiments, showcasing its potential for improving model performance.

\subsection{Synthetic SFT Data Generation}
\label{sec:synthetic_data}

In this work, we investigate the creation of synthetic SFT data through a bootstrapping process involving code descriptions generated by LLMs. 
To be precise, we undertake the task of identifying and refining \textit{self-contained} Verilog modules sourced from Github data~\cite{thakur2023benchmarking}. 
Subsequently, we employ the prompt template depicted in~\Cref{fig:machine_description} to generate corresponding descriptive texts for each of these Verilog modules, effectively creating \textbf{machine} descriptions and code pairs. 
It's worth noting that our approach to synthetic data generation is straightforward in its implementation, and we acknowledge the potential for more advanced techniques as a promising future direction.

We leverage Pyverilog~\cite{pyverilog} to extract to abstract syntax tree from Verilog code and employ the following filtering process to identify \textit{self-contained} Verilog modules from open-sourced Github Verilog code~\cite{thakur2023benchmarking}:
\begin{itemize}
    \item We verify that the filtered code contain the \texttt{module} and \texttt{endmodule} keywords, positioned at the beginning and end of the code, respectively.
    \item We remove Verilog modules more than 200 lines of code or exceeding 1024 tokens.
    \item We ascertain that the code includes at least one of the essential keywords: \texttt{always, assign, always\_ff, always\_comb, always\_latch}.
    \item We ensure extracted modules are \textit{self-contained} without any module instantiation.
\end{itemize}

We further perform approximate deduplication based on MinHash algorithm~\cite{minhash} using \textit{Jaccard} similarity threshold of 0.8 as in~\cite{dedupe}.
We used \texttt{gpt-3.5-turbo} to generate code descriptions based on the prompt template in~\Cref{fig:machine_description}, using \textbf{VerilogEval-human} descriptions of \texttt{shift18, rule110, lemmings1, fsm3onehot} as \textit{few-shot} examples.
We selected these examples with the aim of encompassing a wide range of design instances and the utilization of natural language descriptions, including those presented in tabular formats.
In total we generated 8,502 problem description and code pairs.

\subsection{Results on Supervised Fine-tuning}
\label{sec:sft_results}
We conduct extensive experiments on fine-tuning with the generated synthetic SFT data.
Including both description and code, our SFT data is 11MB in file size, compared with $\sim$700MB of Github Verilog data used in~\cite{thakur2023benchmarking}.
For the fine-tuning process, we employed the Adam optimizer with hyperparameters $\beta_1=0.9$, $\beta_2=0.999$, and $\epsilon=10^{-8}$. 
We set the learning rate to $lr=2e^{-5}$, effective batch size as 1M tokens, and opted not to apply weight decay. 
For all our experiments, we sample $n=20$ code completions for measuring $pass@k=\{1,5,10\}$ using~\Cref{eq:pass_k}.
We use nucleus sampling~\cite{nucleus_sampling} with top $p=0.95$, temperature $temp=0.8$, and context length of 2048.
We used a single NVIDIA DGX node with 8 A100s and 2TB RAM.

Our experimentation primarily focuses on the CodeGen model series~\cite{codegen} and its Verilog-trained counterparts in~\cite{thakur2023benchmarking}. 
These experiments encompass model sizes of \texttt{350M}, \texttt{2B}, \texttt{6B}, and \texttt{16B}.
We use \texttt{-sft} to indicate models fine-tuned with our synthetic SFT data.
We clarify our notation for base models as follows:
\begin{itemize}
    \item \texttt{codegen-nl}~\cite{codegen} : Natural language model. Trained on \textrm{ThePile}~\cite{the_pile} 825.18GB English text corpus.
    \item \texttt{codegen-multi}~\cite{codegen}: Code model. Initialized from \texttt{codegen-nl} and continue trained on \textrm{BigQuery} multi-lingual code dataset consisting of C, C++, Go, Java, JavaScript, and Python.
    \item \texttt{codegen-verilog}~\cite{thakur2023benchmarking}: Verilog code model. Initialized from \texttt{codegen-multi} and continue trained on $\sim$300MB of Github Verilog and 400MB of textbook data.
\end{itemize}
Furthermore, we conducted comparisons with the \texttt{gpt-3.5-turbo} and \texttt{gpt-4} models through OpenAI APIs~\cite{openai}. 
Our analysis specifically involved default \texttt{4k} context length models from \texttt{0613} checkpoints.

\begin{figure}[ht]
  \centering
    \subfloat{\includegraphics[width=0.24\textwidth]{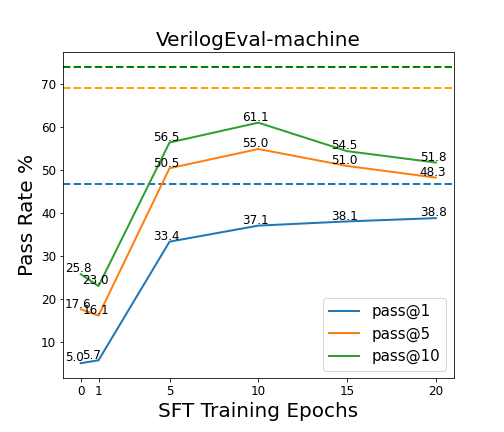 }}
    \subfloat{\includegraphics[width=0.24\textwidth]{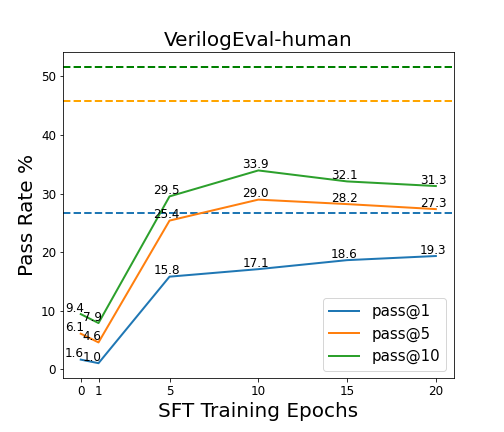}} \\
    {\small (a) \texttt{codegen-16B-multi-sft}\label{fig:sft_epoch_a}}
    \subfloat{\includegraphics[width=0.24\textwidth]{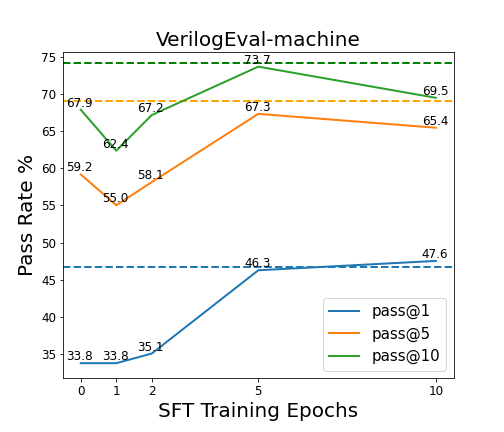 }}
    \subfloat{\includegraphics[width=0.24\textwidth]{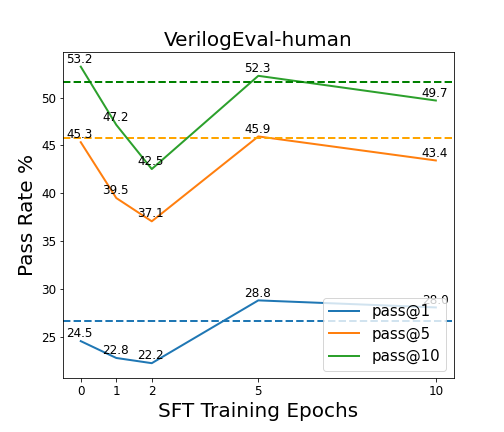}} \\
    {\small (b) \texttt{codegen-16B-verilog-sft}\label{fig:sft_epoch_b}}
    \caption{SFT training epochs and pass rate on \textbf{VerilogEval}. Dashed lines are \texttt{gpt-3.5} results.}\vspace{-0.2in}
    \label{fig:sft_epochs}
\end{figure}

\subsubsection{Training Epochs}
\Cref{fig:sft_epochs} depicts the pass rate on \textbf{VerilogEval} with different SFT training epochs.
Dashed lines indicate \texttt{gpt-3.5} results.
Results show that \textbf{machine} descriptions correlate well with \textbf{human}, demonstrating that synthetic generated benchmarks could be a good indicator for downstream task performance.

In most cases, we observe that the performance metric $pass@1$ continues to exhibit improvement as the supervised fine-tuning (SFT) training epochs progress, whereas the metrics $pass@5$ and $pass@10$ begin to deteriorate. 
This trend suggests that with an increase in training epochs, the model tends to overfit to the SFT data, limiting its ability to generate diverse solutions for tackling complex challenges. 
Interestingly, this overfitting also leads to an increase in the model's confidence and success rate when dealing with simpler problems, highlighting a discernible trade-off between the $pass@1$ and $pass@10$ metrics. 
Consequently, we encourage future research to report on both of these metrics, particularly for models post-alignment, to provide a more comprehensive assessment of their performance.

Throughout the remainder of this study, we conduct supervised fine-tuning (SFT) using 10 epochs for \texttt{multi} and 5 epochs for \texttt{verilog} models. 

\subsubsection{Model Size and Base Model}
\begin{figure}[ht]
  \centering
    \subfloat{\includegraphics[width=0.24\textwidth]{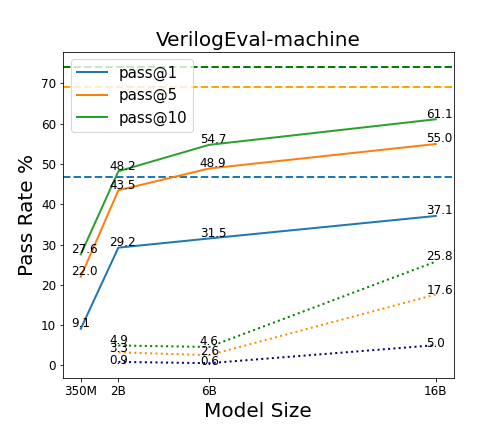}}
    \subfloat{\includegraphics[width=0.24\textwidth]{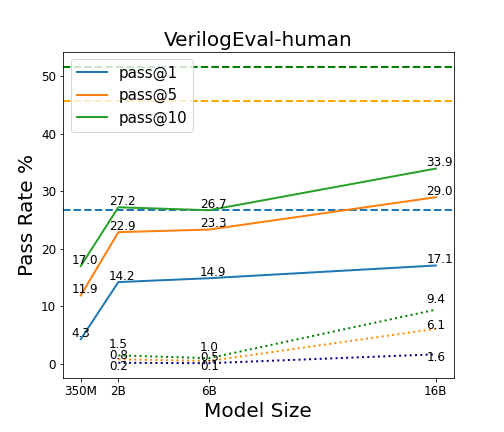}} \\
    {\small (a) \texttt{codegen-multi-sft}\label{fig:sft_size_a}}
    \subfloat{\includegraphics[width=0.24\textwidth]{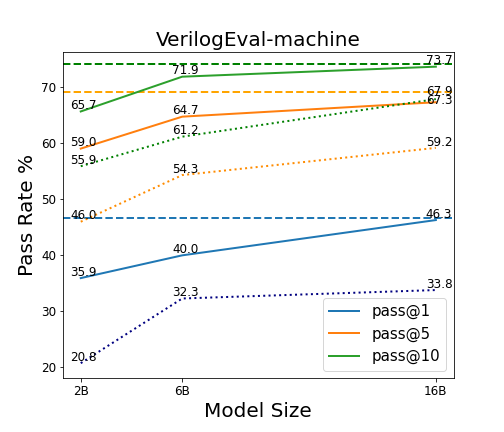}}
    \subfloat{\includegraphics[width=0.24\textwidth]{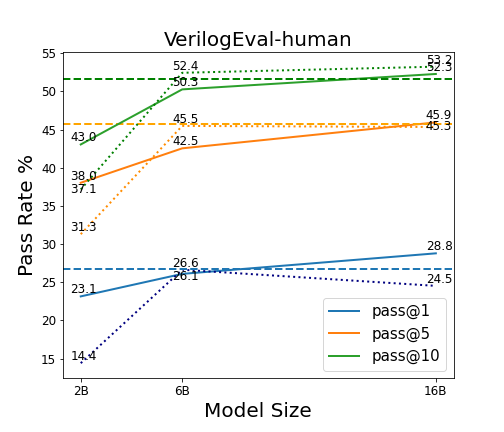}} \\
    {\small (b) \texttt{codegen-verilog-sft}\label{fig:sft_size_b}}
    \caption{\textbf{VerilogEval} results on different model size. Solid lines are \texttt{sft} models, dotted lines are corresponding base models without SFT, dashed lines are \texttt{gpt-3.5} results.}
    \label{fig:sft_model_size}
\end{figure}

\Cref{fig:sft_model_size} illustrates the pass rates for the \textbf{VerilogEval} task using various model sizes and base models. 
The base model denotes the initial model checkpoint prior to SFT. 
It is worth noting that we have omitted the results for models with a size of \texttt{350M}, either due to their unavailability or because their pass rates are insufficient to demonstrate statistical significance.
Our results suggest that more capable and larger models generally result in better Verilog coding capabilities.

In most instances, SFT using synthetically generated data yields notable enhancements in downstream model performance. 
These improvements are particularly pronounced, especially in the case of \texttt{multi} models, where the original model was not explicitly trained on a substantial corpus of Verilog code. 
In the case of \texttt{verilog} models, \textbf{VerilogEval-machine} exhibited significant performance gains, whereas the \textbf{VerilogEval-human} approach displayed comparatively less improvement and, at times, even slight deteriorations. 
Our SFT data is sourced from the GitHub Verilog corpus, and thus does not introduce additional Verilog code that the model did not encounter during its training for \texttt{verilog} models. 
However, by providing problem-code pairs, this data facilitates better alignment of the model, resulting in improved outcomes for \textbf{VerilogEval-machine}.
Despite incorporating \textit{few-shot} prompting during the generation of SFT data (as discussed in ~\Cref{sec:synthetic_data}), the generated descriptions tend to be primarily low-level, lacking the textual diversity found in \textbf{human} examples, such as state transition graphs, waveforms, Karnaugh maps, and similar elements.
This ``mis-alignment'' between SFT data and \textbf{VerilogEval-human} might have caused \texttt{verilog-sft} models to degrade slightly in performance.
We envision that increasing SFT (and Verilog pretraining) data diversity and quality would further lead to increased performance.

\begin{table}[ht]
\centering
\resizebox{0.95\linewidth}{!}{%
\begin{tabular}{c|ccc|ccc}
\hline
\multirow{2}{*}{Model}                                & \multicolumn{3}{c|}{\textbf{VerilogEval-machine}}                                      & \multicolumn{3}{c}{\textbf{VerilogEval-human}}                                        \\ \cline{2-7} 
                                                      & \multicolumn{1}{l}{pass@1} & \multicolumn{1}{l}{pass@5} & \multicolumn{1}{l|}{pass@10} & \multicolumn{1}{l}{pass@1} & \multicolumn{1}{l}{pass@5} & \multicolumn{1}{l}{pass@10} \\ \hline
\texttt{gpt-3.5}                                      & 46.7                       & 69.1                       & 74.1                         & 26.7                       & 45.8                       & 51.7                        \\ \hline
\texttt{gpt-4}                      & 60.0                       & 70.6                       & 73.5                         & 43.5                       & 55.8                       & 58.9                        \\ \hline \hline
\texttt{verilog-sft} & 46.2                       & 67.3                       & 73.7                         & 28.8                       & 45.9                       & 52.3                        \\ \hline
\end{tabular}}
\caption{Results on \texttt{gpt} models, comparing with \texttt{codegen-16B-verilog-sft}.}
\label{tab:gpt_comparision}
\end{table}

In~\Cref{tab:gpt_comparision}\footnote{Our prior version contained errors in numbers reported for \texttt{gpt-4}.} we present the results obtained from both the \texttt{gpt-3.5} and \texttt{gpt-4} models for the \textbf{VerilogEval} task.
Additionally, we demonstrate that our top-performing model \texttt{codegen-16B-verilog-sft}, exhibits performance that is on par with \texttt{gpt-3.5}.

\begin{table}[ht]
\centering
\resizebox{0.8\linewidth}{!}{%
\begin{tabular}{c|ccc}
\hline
\multirow{2}{*}{Model}         & \multicolumn{3}{c}{\textbf{VerilogEval-machine}}                                      \\ \cline{2-4} 
                               & \multicolumn{1}{l}{pass@1} & \multicolumn{1}{l}{pass@5} & \multicolumn{1}{l}{pass@10} \\ \hline
\texttt{codegen-16B-nl-sft}    & 33.9                       & 51.9                       & 58.1                        \\ \hline
\texttt{codegen-16B-multi-sft} & 37.1                       & 55.0                       & 61.1                        \\ \hline
\end{tabular}}
\caption{Comparing \texttt{nl} and \texttt{multi} as SFT base models.}
\label{fig:base_compare}
\end{table}

In~\Cref{fig:base_compare} we present a comparison of results between \texttt{sft} models utilizing two distinct base models: \texttt{codegen-nl} and \texttt{codegen-multi}. 
The tokenizer of \texttt{codegen-nl} model is inefficient in handling whitespaces, consequently preventing some of the \textbf{VerilogEval-human} problems from fitting within the limited context window of \texttt{2048} tokens.
Thus we only display results for \textbf{VerilogEval-machine}. 
Despite the fact that \texttt{multi} models undergo pretraining on an extensive corpus of multi-lingual code data, they exhibit only marginal enhancements of approximately 3\% when applied to Verilog coding task. 
This observation potentially suggests that there is limited positive knowledge transfer between software programming languages like C++ and hardware descriptive languages such as Verilog. 
This highlights the significance of pretraining on substantial Verilog corpora, as it can significantly enhance model performance in Verilog-related tasks.

\subsubsection{SFT Data Quality}
We conducted a comparative experiment aimed at assessing the significance of data quality in SFT. 
In this experiment, we introduced a manipulation by shuffling problem descriptions with incongruous Verilog code solutions, resulting in the creation of erroneous problem-code pairs denoted as \texttt{sft-error}. 
The outcomes, as presented in \Cref{tab:data_quality}, provide a comparison of the performance results obtained through fine-tuning on the \texttt{codegen-2B-verilog} models concerning the \textbf{VerilogEval-machine} task. 
The results clearly demonstrate that the inclusion of incorrect problem-code pairs detrimentally impacts model performance, underscoring the critical importance of maintaining high-quality SFT data.

\begin{table}[ht]
\centering
\resizebox{0.9\linewidth}{!}{%
\begin{tabular}{c|ccc}
\hline
\multirow{2}{*}{Model}                & \multicolumn{3}{c}{\textbf{VerilogEval-machine}} \\ \cline{2-4} 
                                      & pass@1         & pass@5         & pass@10        \\ \hline
\texttt{codegen-2B-verilog}           & 20.1           & 46.0           & 55.9           \\ \hline
\texttt{codegen-2B-verilog-sft}       & 35.9           & 59.0           & 65.7           \\ \hline
\texttt{codegen-2B-verilog-sft-error} & 21.4           & 38.8           & 46.1           \\ \hline
\end{tabular}}
\caption{Comparative experiment on SFT data quality. Incorrect low-quality SFT data degrades model performance.}
\label{tab:data_quality}
\end{table}
\section{Limitations and Future Directions}
In \textbf{VerilogEval}, our primary focus centers on harnessing Large Language Models (LLMs) to generate \textit{self-contained} Verilog modules directly from natural language text descriptions. 
While we incorporate a wide array of hardware design topics through human-generated descriptions, it's important to note that our current evaluations are confined to boilerplate code generation for relatively small-scale designs. 
We emphasize the significance of module instantiation as a crucial capability in Verilog, as it plays a pivotal role in constructing complex system-level designs, albeit currently absent from our benchmark.
Recent advancements in LLM-based coding benchmarking, as seen in~\cite{codereval}, are starting to explore pragmatic code generation beyond standalone functions. 
It's worth mentioning that our testing environment solely assesses functional correctness and does not ensure that the generated Verilog code adheres to synthesizable formatting standards. 
We do not evaluate the performance of downstream circuit implementations, a gap that is addressed by the work presented in~\cite{RTLLM}.

Additionally, it's crucial to recognize that boilerplate Hardware Description Language (HDL) code generation, as currently addressed in our \textbf{VerilogEval} and similar endeavors, inherently operates within an exceedingly limited scope within the broader landscape of hardware design. 
Hardware design, in its entirety, necessitates a multidisciplinary approach that draws on the expertise of domain professionals ranging from transistor device, circuit design, and to hardware system architecture. 
This holistic understanding is indispensable, as it allows design teams to navigate the intricacies of hardware design effectively.
Furthermore, it's important to highlight that a significant portion of the hardware design process revolves around optimizing the Power, Performance, and Area (PPA) metrics. 
These three factors, power consumption, computational performance, and physical chip area, are paramount considerations in modern hardware design. 
Achieving the right balance among them is a formidable challenge that requires meticulous planning, advanced simulation, and iterative refinement. 
Equally critical is the extensive effort invested in design verification, aimed at ensuring the reliability and yield of the hardware. 
Verifying that a design functions as intended under diverse conditions and corner cases is vital to mitigate the risk of costly errors and to guarantee the final product meets its specifications. 
In essence, the successful realization of hardware designs hinges on the convergence of domain expertise, PPA optimization, and robust verification practices.

Nonetheless, Large Language Models (LLMs) present an exciting opportunity for future research to revolutionize the hardware design process. 
This transformative potential lies in their ability to collaborate with domain experts in formulating novel problems and devising innovative solutions. 
By leveraging the vast knowledge and natural language understanding of LLMs, domain experts can work in tandem with these models to explore uncharted territories in hardware design, potentially leading to breakthroughs that enhance the efficiency, reliability, and agility of the design process. 
The fusion of human expertise with machine intelligence using LLMs in this collaborative endeavor promises an exhilarating avenue for future research, one that holds the potential to reshape the very fabric of the hardware design research landscape.
\section{Conclusion}
\label{sec:conclusions}
The growing prominence of Large Language Models (LLMs) has ushered in a new era of their application across various domains. 
In this paper we introduce a specialized benchmarking framework meticulously designed to assess LLM performance within the realm of Verilog code generation for hardware design. 
The cornerstone of this contribution lies in the creation of a robust evaluation dataset, comprising of 156 distinct problems sourced from HDLBits. 
Furthermore, we have demonstrated that the Verilog code generation capabilities of pretrained language models can be enhanced through supervised fine-tuning, facilitated by the generation of synthetic problem-code pairs using LLMs. 
These findings not only advance the state of the art in Verilog code generation but also underscore the vast potential of LLMs in shaping the future of hardware design and verification.

\section*{Acknowledgment}
The authors would like to thank Henry Wong (henry@stuffedcow.net) the creator of HDLBits for his invaluable assistance in providing reference solutions and testbenches for the problems used in this paper.

\clearpage
\bibliographystyle{IEEEtran}
\bibliography{./ref/Top-sim,./ref/DFM,./ref/HSD,./ref/additional,./ref/PD,./ref/LLM}

\end{document}